\pdfoutput=1

\documentclass[11pt]{article}

\usepackage[final]{acl}

\usepackage{times}
\usepackage{latexsym}

\usepackage{hyperref}
\usepackage{cleveref}
\usepackage{makecell}
\usepackage{array}
\usepackage{booktabs}

\usepackage{rotating}
\usepackage{array,multirow,graphicx}

\usepackage{caption}
\usepackage{subcaption}

\usepackage[T1]{fontenc}

\usepackage[utf8]{inputenc}
\usepackage{float}     
\usepackage{placeins}  

\usepackage{microtype}

\usepackage{inconsolata}

\usepackage{graphicx}

\usepackage{tikz}

\usetikzlibrary{positioning, shapes.multipart}
\newcommand{\voc}[1]{\texttt{#1}}

\title{Generating Text from Uniform Meaning Representation}

\author{Emma Markle \\
  Amherst College \\
  \texttt{emarkle26@amherst.edu} \\\And
  Reihaneh Iranmanesh \\
  Amherst College \\
  \texttt{riranmanesh25@amherst.edu} \\\And
  Shira Wein \\
  Amherst College \\
  \texttt{swein@amherst.edu} \\}

\begin{document}
\maketitle
\begin{abstract}
Uniform Meaning Representation (UMR) is a recently developed graph-based semantic representation, which expands on Abstract Meaning Representation (AMR) in a number of ways, in particular through the inclusion of document-level information and multilingual flexibility. In order to effectively adopt and leverage UMR for downstream tasks, efforts must be placed toward developing a UMR technological ecosystem.
Though only a small amount of UMR annotations have been produced to date, in this work, we investigate the first approaches to producing text from multilingual UMR graphs. Exploiting the structural similarity between UMR and AMR graphs and the wide availability of AMR technologies, we introduce (1)~a baseline approach which passes UMR graphs to AMR-to-text generation models,
(2)~a pipeline conversion of UMR to AMR, then using AMR-to-text generation models, and (3)~a fine-tuning approach for both foundation models and AMR-to-text generation models with UMR data.
Our best performing models achieve multilingual BERTscores of 0.825 for English and 0.882 for Chinese,  a promising indication of the effectiveness of fine-tuning approaches for UMR-to-text generation even with limited UMR data.\footnote{Our checkpoints and code are available at \url{https://github.com/ACNLPlab/UMR-Text-Gen}{}}

\end{abstract}

\section{Introduction}

\begin{figure}[t]
    \small
\centering
\resizebox{\linewidth}{!}{
\begin{tikzpicture}[
blue/.style={rectangle, draw=black, very thick, minimum size=6mm},
]
	\node[blue](s) at (10,10) {\voc{search-01}};
	\node[blue](p) at (7,8) {\voc{person}};
    \node[blue](t) at (6,5) {\voc{3rd}};
    \node[blue](sing1) at (8,5) {\voc{Singular}};
	\node[blue](a) at (9,8) {\voc{Activity}};
	\node[blue](f) at (11,8) {\voc{FullAff}};
    \node[blue](c) at (13,8) {\voc{clue}};
    \node[blue](sing2) at (13,5) {\voc{Singular}};
	\draw[->, thick] (s.south) -- (p.north) node[midway, above, sloped] {\voc{:ARG0}};
	\draw[->, thick] (s.south) -- (c.north) node[midway, above, sloped] {\voc{:ARG1}};
 	\draw[->, thick] (s.south) -- (a.north) node[pos=0.7, sloped, above] {\voc{:aspect}};
 	\draw[->, thick] (s.south) -- (f.north) node[pos=0.65, sloped, above] {\voc{:modstr}};
    \draw[->, thick] (c.south) -- (sing2.north) node[midway, above, sloped] {\voc{:refer-number}};
    \draw[->, thick] (p.south) -- (sing1.north) node[midway, above, sloped] {\voc{:refer-number}};
    \draw[->, thick] (p.south) -- (t.north) node[midway, above, sloped] {\voc{:refer-person}};

\end{tikzpicture}}
\smallbreak
\small
\begin{verbatim}
(s / search-01
  :Arg0 (p / person
    :refer-person 3rd
    :refer-number Singular)
  :Arg1 (c / clue
    :refer-number Singular)
  :aspect Activity
  :modstr FullAff)
\end{verbatim}
\caption{UMR graph for the sentence ``He was searching for a clue'' in graph form (top) and in `PENMAN' notation \citep{kasper-1989-flexible} (bottom).
}
\label{fig:umr_example}
\end{figure}

Uniform Meaning Representation (UMR) is a graph-based semantic representation designed to capture the core elements of meaning for a wide range of languages, accounting for some languages' limited resources and linguistic diversity in the annotation process \citep{van2021designing}. An example sentence-level UMR graph can be seen in \Cref{fig:umr_example}.
The UMR project has seen recent progress but is still in the early stages, with new resources such as an online web annotation tool \citep{ge-etal-2023-umr} and a relatively small recently-released annotated dataset \citep{bonn-etal-2024-building-broad}.

Abstract Meaning Representation \citep[AMR;][]{banarescu-etal-2013-abstract},
on which UMR is based, is an annotation schema designed for English which has seen success and adoption by the broader NLP community, including incorporation into downstream applications and studies of language \citep{wein-opitz-2024-survey}; this is in large part due to the substantial efforts made towards high-quality text-to-AMR parsing and AMR-to-text generation models \citep{sadeddine-etal-2024-survey}.\footnote{The breadth of parsing and generation work for AMR can be seen in the \href{https://nert-nlp.github.io/AMR-Bibliography/}{AMR Bibliography}.}
Thus, in order to see similar success for the multilingual UMR schema, efforts towards UMR-to-text generation and text-to-UMR parsing are critical. 
As such, the multilingual (and document-level) benefits of UMR cannot be exploited by the NLP community until the tools exist to easily parse and generate.

In this work, we tackle the task of UMR-to-text generation and introduce the first models for this task.
Our experimentation focuses on primarily English and Chinese, but leverages data in all of the six languages included in the UMR v1.0 dataset (detailed in \Cref{ssec:data}): English, Chinese, Sanapaná, Arápaho, Kukama, and Navajo.
We introduce three approaches for this task, all of which account for the limited amount of UMR resources by exploiting the breadth of AMR-to-text generation tools, given the substantial shared features and structure between AMR and UMR. Our approaches and contributions include:
\begin{enumerate}
\addtolength\itemsep{-3mm}
    \item A baseline analysis as to how well six AMR-to-text generation models generate text from UMR out-of-the-box (\Cref{sec:baseline}).
    \item A novel pipeline approach to UMR-to-text generation, which converts UMR graphs into AMRs, then uses them as input to AMR-to-text generation models (\Cref{sec:conversion}).
    \item 
    Seven fine-tuned UMR-to-text generation models, using three  pretrained large language models and four pretrained AMR-to-text generation models as foundations
    (\Cref{sec:fine-tuned}). 
\end{enumerate}
Our experiments reveal that while our baseline approach of using UMRs as input to AMR-to-text models does produce text, it often fails to handle UMR-specific terminology.
Our fine-tuned models, particularly models pretrained on AMR data, achieve both high adequacy and fluency, and in fact are able to leverage the document-level information contained in UMR graphs for better output despite only generating a single sentence at a time. Additionally, the best performing models are fine-tuned exclusively on the same language UMR data, suggesting that UMR data in other languages may not be helpful in enhancing monolingual UMR-to-text generation models.

\section{Uniform Meaning Representation}

UMR is based on Abstract Meaning Representation (AMR), 
which was designed for English 
\citep{banarescu-etal-2013-abstract} but has since seen various individual language extensions and cross-lingual applications \citep{10.1162/coli_a_00503}. 
While AMR requires these individual language-specific adaptations to  accommodate the linguistic features of the individual language (such as pronoun drop, grammatical number, and affixes), UMR's flexibility enables annotation of many languages within the same schema, showing its promise as a multilingual representation \citep{wein-bonn-2023-comparing}.

Similarly to AMR, UMR annotations are rooted, directed graphs that capture meaning \citep{van2021designing}.
Unlike AMR,
UMR additionally incorporates aspect and modality at the sentence-level, contains document-level information (enabling annotation of coreferential relations), and alignments between the coreferential elements. 

In order to ensure that UMR could reflect meaning for many languages, the UMR schema is flexible in its annotation while ensuring consistency across languages, differentiating it from the English-based AMR.
At the sentence-level, UMR accounts for linguistic diversity across languages through the use of a lattice-like annotation schema \citep{van-gysel-etal-2019-cross}, enabling annotators to apply a more coarse-grained or fine-grained annotation 
based on the features of the language being annotated.
Particular care is also given to the annotation of low- or no-resource languages \citep{vigus-etal-2020-cross}, as ``Stage 0'' annotation is performed for languages that do not have existing rolesets available. Stage 0 annotation enables annotators to establish predicate-argument structures and develop rolesets while performing UMR annotation.

While text-to-AMR parsers and AMR-to-text generation models have seen substantial progress in recent years (with Smatch scores of 86\% for parsing \citep{lee-etal-2022-maximum,vasylenko-etal-2023-incorporating} and 50 BLEU points for generation \citep{cheng-etal-2022-bibl}), no efforts have yet been made towards UMR-to-text generation models, largely due to the recent development of the schema.
We opt to use AMR-to-text models as a baseline for UMR-to-text generation, though the fact that UMR contains \emph{more} semantic information than AMR (such as aspect, modality, and document-level information) makes AMR-to-text models not particularly well-suited for this task.
Models trained solely on AMR are unlikely to correctly account for these elements, leading to a loss of critical semantic information.
The use of AMR-to-text generation models as a baseline motivates progress on dedicated UMR-to-text generation approaches, such as ours.


Recent work has investigated
automated Arápaho UMR annotation \citep{buchholz-etal-2024-bootstrapping-umr}, automatic annotation of tense and aspect for UMR \citep{chen-etal-2021-autoaspect}, and conversion of AMR graphs to UMR \citep{bonn-etal-2023-mapping,post-etal-2024-accelerating}.
Special attention has been given to UMR annotation of multi-word expressions \citep{bonn-etal-2023-umr} and Chinese verb compounds \citep{sun-etal-2023-umr}.

Contributing to the UMR technological ecosystem, \Citet{chun-xue-2024-pipeline} released a text-to-UMR parser, which produces the document-level UMR graph based on the contents of the sentence, and the sentence-level UMR graph by running existing text-to-AMR parsers and then converting the AMR into a UMR. This pipeline parsing approach mirrors our baseline approach, in the reverse direction.


\section{Methodology}
Given that in this work we are developing the first approaches to UMR-to-text generation, here we first establish the data (\Cref{ssec:data}) and evaluation methods (\Cref{ssec:evaluation,ssec:indigenous,ssec:human-eval}) that we use for this task.
Then, in the sections that follow, we introduce each of our three different approaches to text generation, which all exploit the wealth of AMR technologies in different ways.

First, our baseline approach uses six AMR-to-text generation models, passing the sentence-level UMR graphs as input (\Cref{sec:baseline}).
Next, using the same six models used for our baseline, we develop a pipeline approach (\Cref{sec:conversion}) to UMR-to-text generation, which involves first converting the UMR graphs to AMR graphs, and then passing the converted AMR graphs as input to the models. 
Then, based on the baseline performance of the AMR-to-text generation models, we fine-tune the top performing models as well as three large language models (with no AMR input) on the UMR training set (\Cref{sec:fine-tuned}).

\subsection{Data}
\label{ssec:data}
We use the first release of UMR data \citep{bonn-etal-2024-building-broad}, which contains annotations in English, Chinese, and four languages indigenous to the Americas: Arápaho, Navajo, Sanapaná, and Kukama. The English data contains
LORELEI news text and a description of a silent film. The Chinese data consists of wikinews sentences.
The Arápaho, Navajo, and Kukama annotations all represent narrative documents, while it is not clear what genre the Sanapaná annotations are \citep{bonn-etal-2024-building-broad}.
We first take out the English UMR data that contains equivalent AMRs in the AMR3.0 dataset \citep{knight_amr_3.0}, as to avoid leakage and unfair evaluations of models which contain the corresponding AMR data, removing 66 English UMRs. 

Not all annotations contain both sentence-level and document-level graphs. Henceforth, given that all sentences which have document-level annotations contain both sentence- and document-level graphs, we use \emph{sentence-level annotations} to refer to just the sentence-level graph, and \emph{document-level annotations} to refer to the combination of a sentence-level graph and document-level graph for an individual sentence (note that the document-level graphs contain information about how the sentence in question relates to other sentences in the document).
Our final data split is 70\% for training, 10\% for development, and 20\% for testing (see \Cref{tab:split}). 

\begin{table}
    \hspace{-1cm}
    \small
    \begin{center} 
    \begin{tabular}{|c|c|c|c|}
    \hline
    Language&Training (70\%)&Dev (10\%)&Test (20\%)\\
    \hline
    English& 100 (96)& 13 (13)& 30 (28)\\ 
    \hline
    Chinese& 236 (236)& 40 (40)& 82 (82)\\
    \hline
    Arápaho& 256 (46)& 36 (7)& 114 (54)\\
    \hline
    Navajo& 371 (148)& 52 (20)& 83 (0)\\
    \hline
    Sanapaná& 433 (366)& 62 (53)& 107 (104)\\
    \hline
    Kukama& 76 (76)& 10 (10)& 19 (0)\\
    \hline
    Total&  1472 (968)& 213 (143)& 435 (268)\\
    \hline
    \end{tabular}
    \end{center}
    \caption{Our training, development, and test splits for the UMR data at the sentence-level, with the amount of sentence-level annotations that also had document-level information displayed in parentheses. We only count data as being document-level if it contains both alignment and a document-level graph beyond \texttt{(s / sentence)}.}
    \label{tab:split}
\end{table}

\subsection{Automatic Evaluation}
\label{ssec:evaluation}
We evaluate the generated text via automatic metrics as well as human evaluation.

We compare the generated text from each of our approaches against the reference sentences by using BERTscore \citep{zhang2019bertscore}, given its previously evidenced correlation with human judgments for AMR-to-English text generation \citep{manning-etal-2020-human}. We specifically use multilingual BERTscore (mBERTscore), as well as METEOR \citep{banerjee-lavie-2005-meteor} and BLEU \citep{papineni2002bleu} to enable ease of comparison with future work.

\subsection{Initial Indigenous Language Evaluation}
\label{ssec:indigenous}
Our initial experimentation reveals primarily positive quantitative indications as far as the ability of our models to produce text in the four indigenous languages, 
for example, mBERTscores of 0.799 for Navajo, 0.816 for Sanapaná, 0.780 for Arápaho, and 0.673 for Kukama, as generated by an amrlib model fine-tuned on all languages' sentence-level data.
However, given that these languages are extremely low-resource and are not likely to be well-evaluated by automatic metrics,
we consult speakers of Arápaho and Navajo to provide a human evaluation of the generated text.
The Arápaho and Navajo speakers indicate that, while
these fine-tuned models do a fairly good job at imitating the script of the four indigenous languages, the output is nonsensical and ungrammatical. We thus opt against a full human evaluation or automatic evaluation for the four indigenous languages, with the understanding that even our top-performing models output gibberish for these languages. 
This is likely due to their morphological complexity, necessitating additional resources for coherent generation.
We still leverage the indigenous language data in fine-tuning our models (\Cref{sec:fine-tuned}).

\subsection{Human Evaluation}
\label{ssec:human-eval}
In order to validate the quantitative results obtained by the automatic metrics, we perform a human evaluation, here focusing on English and Chinese. Six college students participated in the evaluation of the English and Chinese texts, three native speakers of English and Chinese each.
Each annotator judged fluency and adequacy on a scale from 1-4. Fluency was judged first, without exposure to the reference, and then adequacy was judged in relation to the reference.

There were a total of four surveys: English fluency, English adequacy, Chinese fluency, and Chinese adequacy, each of which contained 25 questions. For English, we chose to exclude the 5 shortest sentences from the 30 English test sentences, which left us with a final set of 25 sentences. For Chinese, we randomly selected 25 sentences, again from the test set.
Each question in the surveys displayed six sentences, five of which are from our top-performing models (across multiple approaches), as well as the reference.
The instructions provided to the human raters for the English fluency and adequacy surveys can be seen in
\Cref{sec:appendix}.




We evaluated the inter-annotator agreement using the Pearson correlation coefficient \citep{Pearson1895}, calculating pairwise agreement for each of the four surveys (Chinese and English, fluency and adequacy for each).
The average correlation coefficient for English fluency was 0.72, for English adequacy was 0.78, for Chinese fluency was 0.55, and for Chinese adequacy was 0.64. This indicates that there is a strong correlation between the annotators' scores, validating the judgments, while the English raters exhibited even greater agreement.\footnote{This is likely due to the generated text being both more fluent and more adequate for English than Chinese (see \Cref{ssec:pipe_results,ssec:fine_results}). 
} 
Note also that all pair-wise correlation coefficients had a statistically significant p-value ($p < 0.05$).

\section{Baseline Approach}
\label{sec:baseline}
Now that we have established our methods for evaluating UMR-to-text generation tools, we are able to start exploring various methods of generation.

First, we leverage the similarity between UMR and AMR graphs and the prevalence of AMR technologies, by using  AMR-to-text generation models
out-of-the-box. This serves as a baseline model, and we only use sentence-level UMR graphs (as AMR does not contain document-level information), 
to see how they perform as a zero-shot approach with no exposure to UMR.
\footnote{We additionally experiment with a prompt-based LLM baseline, which prompts a GPT4o-mini to generate text from a UMR graph. The prompt-based experiment performs similarly to this baseline, under-performing our other approaches.}

\subsection{Models}
We apply this approach to six AMR-to-text generation models: 
\begin{itemize}
\addtolength\itemsep{-3mm}
    \item[1.] amrlib\footnote{ \href{https://github.com/bjascob/amrlib/tree/master}{amrlib GitHub Repository}}: sequence-to-sequence T5 model, trained on AMR3.0 \citep{knight_amr_3.0}
    \item[2.] AMRBART \citep{bai-etal-2022-graph}: BART-based \citep{lewis-etal-2020-bart} model 
    pretrained on English text, AMR graphs from AMR2.0 \citep{knight_amr_2.0} and AMR3.0, 
    along with 200k silver AMRs parsed by SPRING
    \item[3-4.] SPRING2 and SPRING3 \citep{bevilacqua-etal-2021-one}: BART-based sequence-to-sequence model trained on AMR v3.0 that simplifies AMR parsing and generation by treating them as symmetric tasks. 
    SPRING2 and SPRING3 differ in their training datasets: SPRING2 is trained on AMR2.0 while SPRING3 is trained on AMR3.0 
    \item[5.] BiBL \citep{cheng-etal-2022-bibl}: utilizes the architectural framework of SPRING to align AMR graphs and text, in order to share information across the parsing and generation tasks trained on AMR2.0 and AMR3.0
    \item[6.] Smelting \citep{ribeiro-etal-2021-smelting}: trained mT5 \citep{xue2021mt5} on gold English AMR graphs and sentences as well as silver machine-translated sentences
\end{itemize}

AMRBART, amrlib, SPRING2, SPRING3 and BiBL are all trained to generate English text, while Smelting can produce Spanish, Italian, German, and Chinese text.
Thus, we run the first five models on the English UMR graphs (to generate into English), and run Smelting on the Chinese UMR graphs (to generate into Chinese).

\subsection{Results}
\label{ssec:base_results}

The automatic metric results for the baseline models show a promising indication of the utility of AMR-to-text generation tools for UMR-to-text generation.

Out-of-the-box, the English results from the five baseline models (\Cref{tab:auto-eval-english}) all achieve mBERTscores of around 0.7 (ranging from 0.681-0.704).
The Chinese baseline results (\Cref{tab:auto-eval-chinese}) are similarly high, with the baseline Smelting model achieving a mBERTscore of 0.716. The BLEU and METEOR scores are noticeably lower for both languages, to be expected of those metrics.

Our qualitative analysis of the baseline models reveals the common inclusion of UMR-specific terms, such as \texttt{refer-number singular}, \texttt{full-affirmative}, \texttt{umr-unknown}, and \texttt{3rd person}, in the text output. Examples of this can be seen in the following generated sentence:
\noindent \texttt{Full-affirmative, though, the first singular thought that the second singular would not see the apron at first, is a state of `full affirmative.'}
The only UMR graphs that were converted to highly fluent and adequate text via this baseline approach were one-word sentences such as ``ok'' or ``anyway.''
We also noted that the generated output from the UMR graphs tended to be much longer than the reference output, which was likely due to the inclusion of the UMR-specific terms mentioned previously.\footnote{The average sentence length for each model was 7.3 words for the references, 11.7 for amrlib output, 11.8 for BiBL output, 13.6 for AMRBART output, 14.6 for SPRING3 output, and 14.7 for SPRING2,  output.}

\begin{table}[tb!]  
\setlength{\tabcolsep}{5pt}  
\centering
\small
\begin{tabular}{l|c|c|c}
\hline
\textbf{Models} & \textbf{mBERT} & \textbf{BLEU} & \textbf{METEOR} \\
\hline
Baseline amrlib & \textbf{0.704} & \textbf{0.081} & \textbf{0.386} \\
Baseline amrbart & 0.691 & 0.062 & 0.333 \\
Baseline SPRING3 & 0.681 & 0.050 & 0.298 \\
Baseline SPRING2 & 0.698 & 0.068 & 0.357 \\
\emph{Baseline BiBL} & 0.702 & 0.038 & 0.333 \\
\hline
Pipeline amrlib & 0.761 & 0.119 & 0.418 \\
Pipeline amrbart & 0.775 & 0.181 & 0.409 \\
Pipeline SPRING3 & 0.764 & 0.133 & 0.392 \\
Pipeline SPRING2 & 0.774 & \textbf{0.193} & \textbf{0.436} \\
\emph{Pipeline BiBL} & \textbf{0.784} & 0.159 & 0.428 \\
\hline
FT amrlib-Sent-EN & 0.700 & 0.041 & 0.213 \\
FT SPRING2-Sent-EN & \textbf{0.825} & 0.355 & 0.584 \\
\emph{FT BiBL-Sent-EN} & \textbf{0.825} & 0.289 & 0.570 \\
\emph{FT SPRING2-Doc-EN} & 0.824 & \textbf{0.358} & 0.582 \\
FT BiBL-Sent-EN\&ZH & \textbf{0.825} & 0.355 & \textbf{0.601} \\
FT amrlib-Sent-All & 0.751 & 0.057 & 0.383 \\
FT BiBL-Sent-All & 0.800 & 0.265 & 0.520 \\
FT SPRING2-Sent-All & 0.809 & 0.281 & 0.550 \\
\hline
FT mBART-Doc-EN & 0.750 & 0.104 & 0.411 \\
FT mT5-Sent-EN\&ZH & 0.749 & 0.140 & 0.325\\
FT mT5-Sent-All & 0.772 & 0.181 & 0.368 \\
FT Gemma-Sent-All & 0.440 & 0.000 & 0.004 \\
\emph{FT mBART-Sent-All} & 0.767 & 0.146 & \textbf{0.423} \\
FT mT5-Doc-All & \textbf{0.774} & \textbf{0.201} & 0.417 \\
FT mBART-Doc-All & 0.740 & 0.111 & 0.324 \\

\hline
\end{tabular}
\caption{English automatic evaluation results for each of our three approaches. The fine-tuned (FT) models include their model name, whether the model was fine-tuned on sentence- or document-level information, and the languages included the fine-tuning data. 
Italicized entries were selected for the human evaluation. The fine-tuned models above the third horizontal line are AMR-to-text generation models, while the models below the line are foundation models. Due to having trained over 50 individual models, here we show a representative sample of the best performers from each approach. 
}
\label{tab:auto-eval-english}
\end{table}

\begin{table}[tb!]
\setlength{\tabcolsep}{5pt}
\centering
\small
\begin{tabular}{l|c|c|c}
\hline
\textbf{Models} & \textbf{mBERT} & \textbf{BLEU} & \textbf{METEOR} \\
\hline
\emph{Baseline Smelting} & 0.716 & 0.000 & 0.029 \\
\hline
\emph{Pipeline Smelting} & 0.767 & 0.000 & 0.019 \\
\hline
FT Smelting-Sent-ZH & 0.689 & 0.000 & 0.027 \\
\emph{FT BiBL-Sent-ZH} & 0.881 & 0.247 & 0.593 \\
FT BiBL-Doc-ZH & 0.881 & 0.247 & 0.593 \\
\emph{FT SPRING2-Doc-ZH} & \textbf{0.882} & 0.231 & 0.586 \\
FT BiBL-Sent-EN\&ZH & 0.879 & \textbf{0.250} & \textbf{0.602} \\
FT BiBL-Sent-All & 0.878 & 0.235 & 0.592 \\
\hline
FT mT5-Doc-ZH & 0.815 & 0.122 & 0.375 \\
FT mBART-Sent-All & 0.837 & 0.146 & 0.467 \\
\emph{FT mT5-Sent-All} & \textbf{0.853} & \textbf{0.172} & \textbf{0.476} \\
\hline
\end{tabular}
\caption{Chinese automatic evaluation results for each of our three generation approaches.
Again we show a representative sample of the best performers from each approach. 
}
\label{tab:auto-eval-chinese}
\end{table}

Based on the shorter output length and higher perceived fluency in our initial qualitative analysis (in comparison with the other baseline models, including the higher-scoring amrlib baseline), we determined that BiBL was the best baseline English model, and Smelting was our only baseline Chinese model. 
Thus, we included the baseline BiBL and Smelting models in the human evaluation survey.
The human evaluation scores (\Cref{tab:human-eval-english} and \Cref{tab:human-chinese}) are low with regard to both fluency and adequacy, in spite of the fairly high mBERTscore values. Baseline BiBL for English has a fluency score of 1.44 (scale of 1-4) and an adequacy score of 1.37, while baseline Smelting for Chinese receives a fluency score of 1.69 and an adequacy score of 1.48. The English and Chinese reference sentences receive a fluency score of 3.35 and 3.27, respectively, showing that even the ground-truth sentences are not perceived as perfectly fluent by the annotators, perhaps due to the narrative structure of the UMR data. 
These low scores highlight the inadequacy of the baseline approach, accentuating the need for more advanced techniques to effectively generate text from UMR graphs. 

\section{Pipeline Approach}
\label{sec:conversion}

Our next approach again leverages existing AMR-to-text generation models, but in this case first converting the UMR graphs into AMR graphs. We use all the same models as described in \Cref{sec:baseline}.

\subsection{Conversion Process}
In order to convert UMRs into AMRs, we design a rule-based conversion process.
During conversion, some elements of UMR do not appear in AMR (such as aspect and mode) and are thus simply removed.
Other changes include converting split roles, renaming roles, and introducing additional roles. This process is informed by \citet{bonn-etal-2023-mapping} as well as the AMR \citep{AMRGuidelines} and UMR guidelines
\citep{UMRGuidelines}.
The ``person'' concepts (\texttt{:refer-number} or \texttt{refer-person}) and pronouns generally are handled in a more complicated way in UMR than AMR, so in order to convert these concepts into their AMR counterparts, we changed all ``person'' concepts to a single node of the equivalent English pronoun.
We convert third-person pronouns, both singular and plural, to ``they,'' 
opting for a gender-neutral approach.
Additionally, if a graph refers to the first person but did not include a refer-number tag, we default to using the singular pronoun ``I.''
The Sanapaná data also includes a number of instances of \texttt{:wiki}, which we remove.
In total, this process results in 4 kinds of UMR roles being removed and over 65 kinds of roles converted to their equivalent AMR role.\footnote{We also map \texttt{:material} to \texttt{:consist-of}, \texttt{:concessive-condition} to \texttt{:condition}, and \texttt{have-91} to \texttt{have-03}.}


In order to verify the validity of the conversion process itself, we leverage the 66 held-out English UMRs which have equivalent AMRs in the AMR3.0 dataset (detailed in \Cref{ssec:data}), and compare the generated graphs to their equivalent human-annotated AMR graphs using SMATCH. The average Smatch score for our converter is 0.63, indicating a fairly accurate conversion process. One of the ways in which our rule-based approach to conversion struggles to meet AMR norms is that substantial additional information is contained in the UMR, which is not removed during conversion, such as in the case of the sentence ``Pleasure'' (see \Cref{fig:pleasure_example}).

\begin{figure}[tbh]
    \small
\smallbreak
\small
\noindent Original AMR: 
\begin{small}
\vspace{-1mm}
\begin{verbatim}
(p / pleasure)
\end{verbatim}
\end{small}
\vspace{1mm}
\noindent Original UMR: 
\begin{small}
\vspace{-1mm}
\begin{verbatim}
(s29s / say-01
    :ARG0 (s29p / person)
    :ARG1 (s29h / have-experience-91
        :ARG1 s29p
        :ARG2 (s29p3 / pleasure)
        :ARG3 (s29t / thing)
        :aspect state)
    :ARG2 (s29p2 / person)
    :aspect performance)
\end{verbatim}
\end{small}
\noindent Converted AMR: 
\begin{small}
\vspace{-1mm}
\begin{verbatim}
(s29s / say-01 
    :ARG0 (s29p / person) 
    :ARG1 (s29h / have-experience-91 
        :ARG1 s29p 
        :ARG2 (s29p3 / pleasure) 
        :ARG3 (s29t / thing)) 
    :ARG2 (s29p2 / person))
\end{verbatim}
\end{small}

\noindent Generated text from BiBL of the converted AMR: ``People said it was a pleasurable experience.'' \\
\caption{AMR and UMR graphs for the sentence ``Pleasure,'' compared with the AMR graph produced by our UMR-to-AMR pipeline.}
\label{fig:pleasure_example}
\end{figure}


    

\begin{table}[t]
\centering
\small
\begin{tabular}{l|c|c|c}
\hline
\textbf{Models} & \textbf{Fluency} & \textbf{Adequacy} & \textbf{Sum} \\
\hline
Reference & 3.35 & / & 3.35 \\
Baseline BiBL & 1.44 & 1.37 & 2.81 \\
Pipeline BiBL & \textbf{3.59} & 2.71 & 6.30 \\
FT BiBL-Sent-EN & 3.37 & 3.19 & 6.56 \\
FT SPRING2-Doc-EN & 3.44 & \textbf{3.29} & \textbf{6.73} \\
FT mBART-Sent-All & 2.77 & 2.28 & 5.05 \\
\hline
\end{tabular}
\caption{English human evaluation results for the most promising models of each approach.}
\label{tab:human-eval-english}
\end{table}

\begin{table}[t]
\centering
\small
\begin{tabular}{l|c|c|c}
\hline
\textbf{Models} & \textbf{Fluency} & \textbf{Adequacy} & \textbf{Sum} \\
\hline
Reference & \textbf{3.27} & / & 3.27 \\
Baseline Smelting & 1.69 & 1.48 & 3.17 \\
Pipeline Smelting & 2.19 & 1.69 & 3.88 \\
FT BiBL-Sent-ZH & 1.76 & 2.79 & 4.55 \\
FT SPRING2-Doc-ZH & 1.92 & \textbf{2.81} & \textbf{4.73} \\
FT mT5-Sent-All & 2.28 & 2.29 & 4.57 \\
\hline
\end{tabular}
\caption{Chinese human evaluation results for the most promising models of each approach.}
\label{tab:human-chinese}
\end{table}

\subsection{Results}
\label{ssec:pipe_results}

Our pipeline approach outperforms the baseline approach for both English and Chinese.
For English, BiBL achieves the highest mBERTscore, 0.784, and
for Chinese, Smelting achieves an mBERTscore of 0.767.


This approach proves to be very effective in reducing the amount of UMR terms that appear in the output, leading to more comprehensible sentences and generally clear output. For example, BiBL's pipeline conversion produces sentences such as: ``The scene is opened by you seeing a tree,'' ``they dumped all their pears into a basket,'' and ``they wear apron-like thing with a huge pocket.''
These examples of generated text illustrate the pipeline's strength at producing fluent text; however, as seen in the examples, the pipeline is not able to accurately generate the applicable pronoun and thus always defaults to ``they.''

The perceived fluency of this approach is validated by the human evaluation survey. For English, this approach leads to highly fluent output, with a score of 3.59 and an adequacy score of 2.71, a substantial improvement over the baseline approach. 
The Chinese also has increased scores of 2.19 for fluency and 1.69 for adequacy. It is evident that this approach is more successful for English, but we do observe quantitative and qualitative improvements for both languages.

While the pipeline approach leverages AMR graphs and the well-developed ecosystem of AMR technologies, it is important to note that UMR contains more linguistics information than AMR. Converting from UMR to AMR therefore inevitably leads to information loss, which can affect the accuracy of generated text with respect to UMR-specific content, such as tense, aspect, and referential features.

An illustrative example concerns pronouns. In UMR, pronouns are richly annotated with attributes such as \texttt{:ref-person} (1st, 2nd, 3rd) and \texttt{:ref-number} (singular, dual, paucal, plural), which capture verbal cross-referencing, overt nominal number, and explicit and implicit arguments, supporting nuanced reference across sentences. In contrast, AMR typically represents pronouns using only surface forms such as ``he,'' ``she,'', or ``they,'' discarding much of the underlying information. 
As a result, converting UMR graphs to AMR inherently loses information encoded in UMR, reducing alignment with the original graph and obscuring distinctions in number, person, or implicit arguments.

This mismatch helps explain why our method of first converting UMR to AMR is unlikely to perform optimally: the intermediate AMR representation cannot carry all the distinctions encoded in UMR, leading to systematic gaps in the generated output.

\section{Fine-tuning Processes}
\label{sec:fine-tuned}

Our final approach involves fine-tuning models with UMR data. We fine-tune two kinds of models: AMR-to-text generation models, and foundation models, which have not been exposed to any AMR previously. In this approach we incorporate the document-level graphs as well as the sentence-level graphs and fine-tune on different combinations of the language data.

\subsection{Methods}
We determine which AMR-to-text generation models to fine-tune based on the results from \Cref{sec:baseline,sec:conversion}, picking the best-performing models for the baseline and pipeline approaches.
Accordingly, the first model we fine-tune is amrlib, which also
follows \citet{wein-2022-spanish} (which fine-tunes the t5wtense AMR-to-English generation model for Spanish text generation). 
Next, we fine-tune SPRING2 and Smelting, given their quantitative performances for English and Chinese, respectively.
Finally, we fine-tune BiBL due to its high scores, relative brevity, and perceived fluency 
(as discussed in \Cref{ssec:base_results}).

Then, in order to investigate whether the inclusion of AMR data in the fine-tuning process aids model performance, we fine-tune large language models with no prior exposure to AMR (as opposed to fine-tuning the AMR-to-text generation models).\footnote{Granted, it is possible that these models have been exposed to AMR graphs in pretraining, though prior work has indicated that foundation models are unable to parse AMR graphs out-of-the-box \citep{ettinger-etal-2023-expert}.} These models are mT5 \citep{xue2021mt5}, mBART \citep{liu2020multilingual}, and Gemma 2B \citep{gemma2024improving}.

We then fine-tune each model using the UMR data. 
In order to determine whether document-level information and UMR data of other languages benefit UMR-to-text generation, we create eight datasets by forming all combinations of sentence or document-level data across English, Chinese, and all languages.
This culminates in 8 fine-tuned versions of each of the 7 models; we highlight the key findings in \Cref{tab:auto-eval-english}.\footnote{Recall that the document-level annotations subsume individual sentence-level annotations as well.}

For reproducibility, details of our hyperparameters follow.
We fine-tune BiBL and SPRING, for 30 epochs with a constant learning rate of 0.0001, incorporating gradient clipping at 2.5 and dropout regularization at 0.25. 
We fine-tune mBART-large-50 and mT5 models for 15 epochs
Similarly, we fine-tune Gemma2-2b using the same optimization parameters.
 
\subsection{Results}
\label{ssec:fine_results}

While the fine-tuning results show more variance than the baseline and pipeline scores, the fine-tuned models consistently outperform those approaches in terms of automatic metrics, fluency, and adequacy.
For English, BiBL and SPRING2 trained on sentence-level English data achieved the highest mBERTSCORE of 0.825. As for Chinese, SPRING2 trained on Chinese document-level data achieved the highest mBERTSCORE of 0.882. Furthermore, these high automatic scores are corroborated by human evaluation results, which show high levels of fluency and adequacy scores for both languages. For English, SPRING2 trained on English document-level data, received the second-highest English fluency rating along with the highest adequacy rating. For Chinese, the best fluency score came from mT5 trained on all languages' sentence-level data, while SPRING2 led in adequacy when trained on Chinese document-level data.

These results point to several broad patterns: (1)~the benefit of document-level data, (2)~the relevance of AMR for UMR technologies, and (3)~the strength of our fine-tuning approaches even with limited UMR annotations. 
First, the inclusion of document-level data appears beneficial, especially for SPRING2. 
This suggests that document-level information helps in the task of generation, even when only producing individual sentences. 

Second, our fine-tuned AMR-to-text models outperform the fine-tuned foundation models such as mT5 and mBART, indicating that models pretrained on AMR carry over information that is beneficial to the generation of UMR graphs, despite UMR’s broader scope of information.

Third, the AMR-to-text fine-tuned models generally perform better at a specific language when they are only trained on that language's data versus being trained on all languages. For example, English generation quality drops slightly when trained on multilingual corpora compared to English-only data. This may be due to the ``curse of multilinguality:'' exposure to additional languages can degrade monolingual abilities \citep{conneau-etal-2020-unsupervised}. Chinese models in particular exhibit more variability and, on average, perform slightly below their English counterparts across automatic and human metrics. This may reflect lower-quality annotations or differences in UMR complexity.

Through qualitative analysis, we discover that different models had specific tendencies during generation. For example, BiBL reproduces ellipses and pauses that can be found in the training data due to the data being narrative annotations. This suggests that BiBL closely mirrors the training input structure, even when it is potentially reducing fluency. Furthermore, SPRING2 produces smoother and more fluent output, while still reflecting the aforementioned narrative cues. While fine-tuned LLMs such as mT5 and mBART showed improvements after fine-tuning, they were consistently outperformed by AMR-to-text generation models. This reveals that the inclusion of AMR data in the training process improves model performance as opposed to solely UMR data.

To illustrate these trends, below are examples of output from mBART and SPRING2 from the same graphs. Our mBART model fine-tuned on sentence-level data of all languages' UMR data produces ``A--nd.. one of the pears.. drops down the floor,'' and ``and he's dumping in the basket with all pears.'' On the other hand, SPRING2 trained on document-level English UMR data produces the much more fluent ``One of the pears.. drops down to the floor,'' and
``and he dumps all his pears into a basket.''
Compared to mBART, SPRING2 produces more structured, grammatically correct, and fluent output, likely due to previous training on AMR data.

In a linguistic error analysis, we find that model failures fall into a few consistent observable categories.
First, pronoun consistency degrades across models, with BiBL and mBART incorrectly switching between pronouns when referring to the same entity.
Furthermore, grammatical and lexical errors appear in otherwise fluent outputs, most visibly in mBART (such as ``one of the pears plucks into the tree''). 
Lastly, models often changed how events were originally framed. For example, in a sentence where the speaker expresses some uncertainty (like ``I don't think you see the apron at first''), our models frequently reframe these as confident statements. This indicates that while models can extract the general story from the UMR graphs, they struggle to preserve the speaker's tone and intent. 

Overall, our fine-tuned AMR-to-text models outperform all other approaches in this work, with the most successful models trained solely on the target language with exposure to both sentence-level and document-level data.



\section{Conclusion}

In this work, we develop tools to generate text from Uniform Meaning Representation graphs, introducing three distinct approaches that make use of existing Abstract Meaning Representation technologies to varying degrees.

Our results indicate that the fine-tuned AMR-to-text generation models are stronger for both English and Chinese.
Further, the best performing models are fine-tuned exclusively on the same language UMR data, suggesting that UMR data in other languages may not be helpful in enhancing monolingual UMR-to-text generation models.
We also find that, while we only generate a single sentence at a time, the document-level information improves model performance.
Finally, we find that leveraging AMR data also benefits UMR-to-text generation in a fine-tuned setting, as the AMR-to-text generation models perform better text generation than the foundation models when both are fine-tuned on UMR data.

Future work may explore advancements toward producing text in other languages from UMR, perhaps leveraging additional external resources.
The annotation and release of additional UMR data will support this effort, as well as lead to improved performance for English and Chinese generation.

\section*{Limitations}

One of our limitations is the small size of the UMRv1.0 dataset, which constrains our model fine-tuning capabilities---especially for low-resource languages where training data is scarce. 
Additionally, we lack robust automatic evaluation metrics for the four indigenous languages, with initial reviews by native speakers indicating the generated outputs do not carry meaning. 
Due to inherent randomness in LLM generation, the LLMs used in this work exhibit slight variability across different runs, which may result in occasionally producing varied responses and thus different BLEU, METEOR, and mBERTscores. 

Our approaches use sequence-to-sequence architectures with linearized graphs to leverage existing AMR technologies given limited UMR data. Future work might explore additional architectures such as graph neural networks which directly encode graph structure, or other forms of encoding the UMR without linearization.

\section*{Acknowledgments}
Thank you to the adjudicators for our human evaluation: Crawford Dawson, Yilin Huang, Hank Hsu, Ryan Ji, Megan Li, and Yichen Liu.
Thank you also to Jiyuan Ji and Javier Gutierrez Bach for comments and experimental feedback, and to Lukas Denk and Andrew Cowell for providing judgments on some of the Navajo and Arápaho sentences, respectively.
This work is supported by the Amherst College HPC, which is funded by NSF Award 2117377.

\bibliography{custom}

\appendix
\section{Full Fine-tuning Automatic Metric Results}
\begin{table}[H]
\setlength{\tabcolsep}{5pt}
\centering
\small
\begin{tabular}{l|c|c|c}
\hline
\textbf{Models} & \textbf{mBERT} & \textbf{BLEU} & \textbf{METEOR} \\
\hline
\emph{amrlib} & 0.751 & 0.057 & 0.383 \\
\hline
\emph{SPRING2} & \textbf{0.809} & \textbf{0.281} & \textbf{0.550} \\
\hline
\emph{BiBL} & 0.800 & 0.265 & 0.520 \\
\hline
\emph{Smelting} & 0.589 & 0.000 & 0.070 \\
\hline
\emph{mT5} & 0.772 & 0.181 & 0.368 \\
\hline
\emph{Gemma2} & 0.440 & 0.000 & 0.004 \\
\hline
\emph{mBART} & 0.767 & 0.146 & 0.423 \\
\hline
\end{tabular}
\caption{\textbf{English} automatic evaluation results for the fine-tuning approach \textit{trained on all languages}.}
\label{tab:auto-eval-full-all-langs}
\end{table}

\begin{table}[H]
\setlength{\tabcolsep}{5pt}
\centering
\small
\begin{tabular}{l|c|c|c}
\hline
\textbf{Models} & \textbf{mBERT} & \textbf{BLEU} & \textbf{METEOR} \\
\hline
\emph{amrlib} & 0.780 & 0.144 & 0.400 \\
\hline
\emph{SPRING2} & 0.821 & 0.312 & 0.596 \\
\hline
\emph{BiBL} & \textbf{0.825} & \textbf{0.355} & \textbf{0.601} \\
\hline
\emph{Smelting} & 0.612 & 0.000 & 0.072 \\
\hline
\emph{mT5} & 0.749 & 0.140 & 0.325 \\
\hline
\emph{Gemma2} & 0.506 & 0.001 & 0.031 \\
\hline
\emph{mBART} & 0.744 & 0.079 & 0.341 \\
\hline
\end{tabular}
\caption{\textbf{English} automatic evaluation results for the fine-tuning approach \textit{trained on English and Chinese}.}
\label{tab:auto-eval-full-eng-and-chinese}
\end{table}

\begin{table}[H]
\setlength{\tabcolsep}{5pt}
\centering
\small
\begin{tabular}{l|c|c|c}
\hline
\textbf{Models} & \textbf{mBERT} & \textbf{BLEU} & \textbf{METEOR} \\
\hline
\emph{amrlib} & 0.700 & 0.041 & 0.213 \\
\hline
\emph{SPRING2} & \textbf{0.825} & \textbf{0.355} & \textbf{0.584} \\
\hline
\emph{BiBL} & \textbf{0.825} & 0.289 & 0.570 \\
\hline
\emph{Smelting} & 0.565 & 0.000 & 0.055 \\
\hline
\emph{mT5} & 0.570 & 0.000 & 0.005 \\
\hline
\emph{Gemma2} & 0.506 & 0.001 & 0.031 \\
\hline
\emph{mBART} & 0.744 & 0.111 & 0.356 \\
\hline
\end{tabular}
\caption{\textbf{English} automatic evaluation results for the fine-tuning approach \textit{trained on English}.}
\label{tab:auto-eval-full-eng}
\end{table}


\begin{table}[H]
\setlength{\tabcolsep}{5pt}
\centering
\small
\begin{tabular}{l|c|c|c}
\hline
\textbf{Models} & \textbf{mBERT} & \textbf{BLEU} & \textbf{METEOR} \\
\hline
\emph{amrlib} & 0.033 & 0.000 & 0.001 \\
\hline
\emph{SPRING2} & 0.870 & 0.188 & 0.553 \\
\hline
\emph{BiBL} & \textbf{0.878} & \textbf{0.235} & \textbf{0.592} \\
\hline
\emph{Smelting} & 0.706 & 0.000 & 0.024 \\
\hline
\emph{mT5} & 0.853 & 0.172 & 0.476 \\
\hline
\emph{Gemma2} & 0.449 & 0.000 & 0.002 \\
\hline
\emph{mBART} & 0.837 & 0.146 & 0.467 \\
\hline
\end{tabular}
\caption{\textbf{Chinese} automatic evaluation results for the fine-tuning approach \textit{trained on all languages}.}
\label{tab:auto-eval-full-zho}
\end{table}

\begin{table}[H]
\setlength{\tabcolsep}{5pt}
\centering
\small
\begin{tabular}{l|c|c|c}
\hline
\textbf{Models} & \textbf{mBERT} & \textbf{BLEU} & \textbf{METEOR} \\
\hline
\emph{amrlib} & 0.027 & 0.000 & 0.002 \\
\hline
\emph{SPRING2} & \textbf{0.882} & 0.246 & 0.587 \\
\hline
\emph{BiBL} & 0.879 & \textbf{0.250} & \textbf{0.602} \\
\hline
\emph{Smelting} & 0.685 & 0.000 & 0.028 \\
\hline
\emph{mT5} & 0.826 & 0.147 & 0.421 \\
\hline
\emph{Gemma2} & 0.510 & 0.001 & 0.055 \\
\hline
\emph{mBART} & 0.834 & 0.150 & 0.469 \\
\hline
\end{tabular}
\caption{\textbf{Chinese} automatic evaluation results for the fine-tuning approach \textit{trained on English and Chinese}.}
\label{tab:auto-eval-engzho-zho}
\end{table}

\begin{table}[H]
\setlength{\tabcolsep}{5pt}
\centering
\small
\begin{tabular}{l|c|c|c}
\hline
\textbf{Models} & \textbf{mBERT} & \textbf{BLEU} & \textbf{METEOR} \\
\hline
\emph{amrlib} & 0.039 & 0.000 & 0.003 \\
\hline
\emph{SPRING2} & 0.880 & 0.229 & 0.572 \\
\hline
\emph{BiBL} & \textbf{0.881} & \textbf{0.247} & \textbf{0.593} \\
\hline
\emph{Smelting} & 0.689 & 0.000 & 0.027 \\
\hline
\emph{mT5} & 0.826 & 0.131 & 0.404 \\
\hline
\emph{Gemma2} & 0.520 & 0.002 & 0.064 \\
\hline
\emph{mBART} & 0.837 & 0.148 & 0.463 \\
\hline
\end{tabular}
\caption{\textbf{Chinese} automatic evaluation results for the fine-tuning approach \textit{trained on Chinese}.}
\label{tab:auto-eval-zho-zho}
\end{table}

\section{Instructions for Human Evaluation}
\label{sec:appendix}

\begin{figure}[H] 
    \centering
    \includegraphics[width=\linewidth]{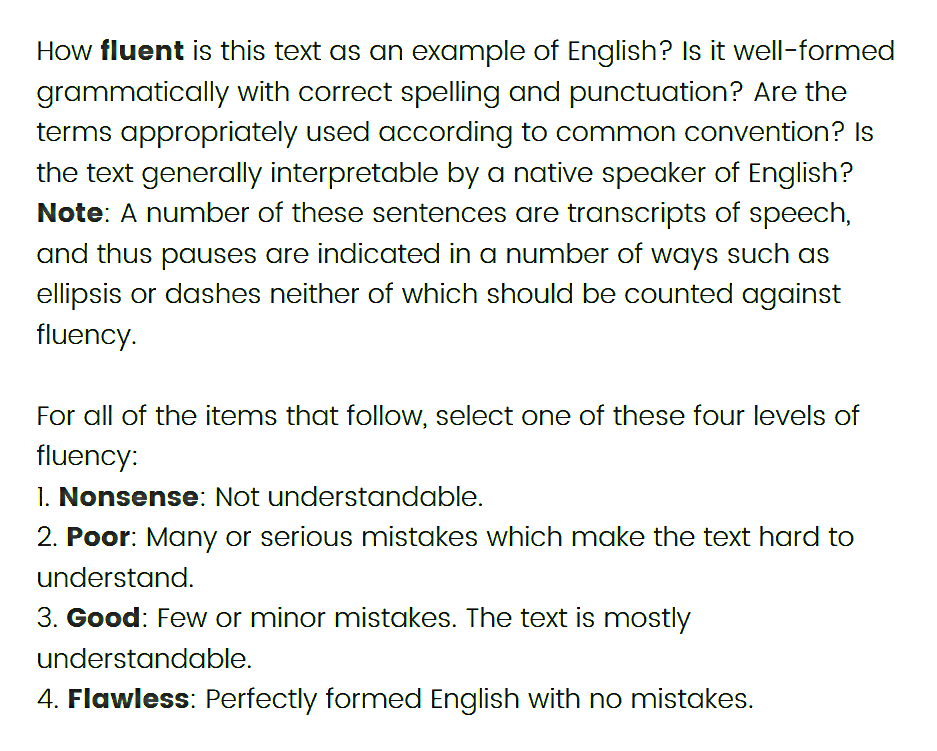}
    \caption{English fluency instructions}
\end{figure}

\begin{figure}[H]
    \centering
    \includegraphics[width=\linewidth]{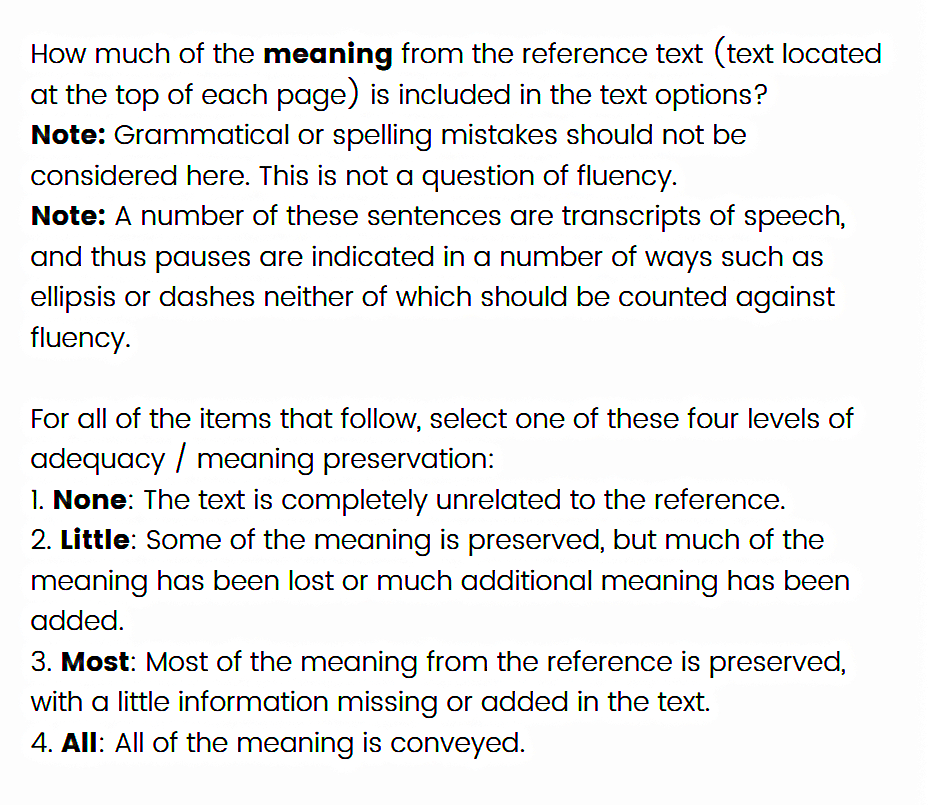}
    \caption{English adequacy instructions}
\end{figure}

\section{Hyperparameters}

\begin{table}[H]
\setlength{\tabcolsep}{4pt}
\centering
\small
\begin{tabular}{l|c|c|c}
\hline
\textbf{Hyperparameter} & \textbf{BiBL/SPRING2} & \textbf{Smelting} & \textbf{Gemma2} \\
\hline
Learning Rate & $1 \times 10^{-4}$ & $1 \times 10^{-4}$ & $1 \times 10^{-4}$ \\
\hline
Epochs & 30 & 15 & 15 \\
\hline
Batch Size (per device) & 8 & 8 & 8 \\
\hline
Gradient Accumulation & 16 & 2 & 2 \\
\hline
Optimizer & -- & AdaFactor & AdaFactor \\
\hline
Max Source Length & 1024 & 1024 & 1024 \\
\hline
Max Target Length & 1024 & 1024 & 2048 \\
\hline
Beam Size & 1 & 5 & 5 \\
\hline
Warmup Steps & 100 & -- & 100 \\
\hline
Weight Decay & 0.004 & -- & 0.004 \\
\hline
Mixed Precision & -- & -- & FP16 \\
\hline
\end{tabular}
\caption{Training hyperparameters for fine-tuning BiBL, SPRING2 (identical to BiBL), Smelting and Gemma2 models.}
\label{tab:hyperparameters}
\end{table}





\end{document}